\documentclass[conference]{ieeeconf}
\usepackage{amsmath,amssymb,amsfonts}
\usepackage{wrapfig}
\usepackage[ruled,vlined,noend]{algorithm2e}
\usepackage{graphicx}
\usepackage{multirow}
\usepackage{tablefootnote}
\usepackage[hidelinks]{hyperref}
\usepackage[capitalise]{cleveref}  
\usepackage[acronym]{glossaries}
\usepackage{glossaries-extra}
\let\labelindent\relax \usepackage{enumitem}
\usepackage{booktabs}
\usepackage{xcolor}
\usepackage{tikz}
\usepackage{xcolor}
\usepackage{setspace}
\usepackage{cite}

\makeglossaries

\newcommand*\circled[1]{\tikz[baseline=(char.base)]{
            \node[shape=circle,draw,inner sep=0.3pt] (char) {#1};}}

\glsdisablehyper
\setabbreviationstyle[acronym]{long-short}
\setabbreviationstyle[short]{short-nolong}
\newacronym{SCLTL}{\textsc{sc}LTL}{Syntactically Co-Safe Linear Temporal Logic}
\newacronym{LTL}{\textsc{ltl}}{linear temporal logic}
\newacronym[category={short}]{LIDAR}{\textsc{lidar}}{\textsc{lidar}}
\newacronym[category={short}]{ProcTHOR}{\textsc{p}roc\textsc{thor}}{\textsc{p}roc\textsc{thor}}
\newacronym{LSP}{\textsc{lsp}}{learning over subgoals planning}
\newacronym{MRLSP}{\textsc{mr-lsp}}{multi-robot learning over subgoals planning}
\newacronym{MCTS}{\textsc{mcts}}{Monte Carlo tree search}
\newacronym{DecPOMDP}{\textsc{d}ec-\textsc{pomdp}}{decentralized partially observable Markov decision process}
\newacronym[longplural=partially observable Markov decision processes]{POMDP}{\textsc{pomdp}}{partially observable Markov decision process}
\newacronym{MDP}{\textsc{mdp}}{Markov decision process}
\newacronym{DFA}{\textsc{dfa}}{deterministic finite automata}
\newacronym[category={short}]{SBERT}{\textsc{sbert}}{\textsc{sbert}}
\newacronym[category={short}]{PDDL}{\textsc{pddl}}{\textsc{pddl}}

\newcommand \Until      {\kern.2em\mathbin{\mathcal{U}\kern.1em}}

\IEEEoverridecommandlockouts

\title{\LARGE \bf Multi-Robot Learning-Informed Task Planning Under Uncertainty}

\author{Abhish Khanal$^{*}$\thanks{* Equal contribution}, Abhishek Paudel$^{*}$, Hung Pham, and Gregory J. Stein%
\thanks{Authors are with the Department of Computer Science,
        George Mason University,
        Fairfax, VA, 22030, USA. Email:
        {\tt\small \{akhanal7, apaudel4, hpham33, gjstein\}@gmu.edu}}%
}

\definecolor{myblue}{rgb}{0.1271,0.4402,0.7075}
\definecolor{myred}{rgb}{0.7925, 0.0932, 0.1129}
\definecolor{mygreen}{rgb}{0.1340, 0.5423, 0.2682}
\begin{document}

\maketitle

\begin{abstract}
We want a multi-robot team to complete complex tasks in minimum time where the locations of task-relevant objects are not known.
Effective task completion requires reasoning over long horizons about the likely locations of task-relevant objects, how individual actions contribute to overall progress, and how to coordinate team efforts.
Planning in this setting is extremely challenging: even when task-relevant information is partially known, coordinating which robot performs which action and when is difficult, and uncertainty introduces a multiplicity of possible outcomes for each action, which further complicates long-horizon decision-making and coordination.
To address this, we propose a multi-robot planning abstraction that integrates learning to estimate uncertain aspects of the environment with model-based planning for long-horizon coordination.
We demonstrate the efficient multi-stage task planning of our approach for 1, 2, and 3 robot teams over competitive baselines in large Proc\textsc{thor} household environments.
Additionally, we demonstrate the effectiveness of our approach with a team of two LoCoBot mobile robots in real household settings.
\end{abstract}

\section{Introduction}
\begin{figure}[t]
    \centering
    \includegraphics[width=\linewidth]{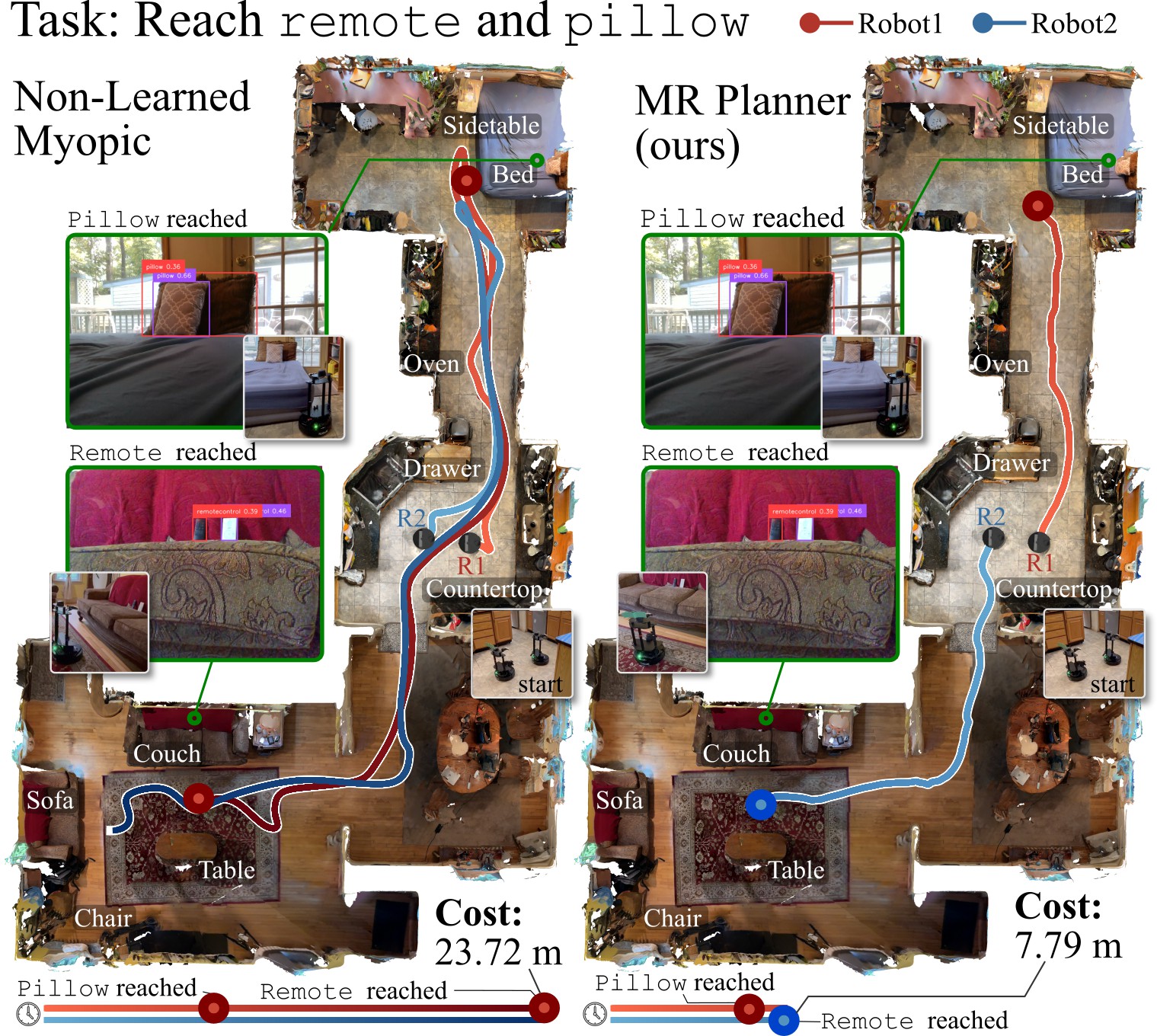}
    \vspace{-2.1em}
    \caption{\textbf{Planner Comparison in Home Environment:} Two robots are tasked to reach \texttt{remote} and \texttt{pillow} with the exact locations of these objects unknown. A myopic approach searches the nearest locations first, until the task is completed, leading to poor behavior. Our approach utilizes learning with a model-based planning framework to best guide the robots for efficient task completion. See further discussion in \cref{sec:real-robot-experiments}.}
    \vspace{-2.2em}
    \label{fig:intro}
\end{figure}

We consider a centrally coordinated team of mobile robots tasked to complete a complex objective in partially known environments in minimum time.
For example, imagine a two-robot team tasked with reaching a remote and a pillow in a home (\cref{fig:intro}).
While the high-level layout of the home is known, the locations of the task-relevant objects are not known.
To complete this task effectively, the robots must coordinate their efforts despite the uncertain duration and outcomes of actions undertaken by members of the team as they search for task-relevant objects.

Effective behavior requires that robots reason over long-horizons about where task-relevant objects are likely to be, how their individual actions contribute to overall task progress, and how to coordinate to most efficiently make progress towards the task.
Planning is an effective tool to reason over long-horizons and achieve such behaviors.

However, planning in this domain is incredibly challenging.
Even when the location of task-relevant objects is known, team coordination requires consideration of which robot does what and when so that they may align their progress to minimize time.
When the locations of task-relevant objects are unknown, planning becomes even more complex, since what the team does next will depend on where and when those objects are found.
Uncertainty results in a multiplicity of possible outcomes, which grows combinatorially as the planning horizon increases, making long-horizon coordination and decision-making extremely challenging.

A naive approach in such scenarios is for each robot to search nearby locations for task-relevant objects.
While simple, this strategy is often inefficient and leads to poor behavior of the team (\cref{fig:intro}, left).
Learning can be used to inform the likely locations of task-relevant objects, yet using this information for effective task coordination still requires long-horizon reasoning among the robots.
As such, good behavior in such scenarios necessiates strategically dividing the search space and effectively coordinating the actions of each robot in the team (\cref{fig:intro}, right).

Achieving such coordinated, long-horizon decision making requires a planning abstraction that
\circled{1} incorporates predictions about what might lie in unseen space (e.g., via learning)
\circled{2} defines a single-robot action abstraction with predicted action outcome probabilities and so understands how each action can advance the overall task, and
\circled{3} leverages this single-robot action abstraction to build a multi-robot state-transition model that supports concurrent execution and coordination among multiple robots.

While no planning abstraction exists that simultaneously supports all such capabilities, progress has been made towards each of these in isolation or in pairs.
There are multiple approaches that develop learning-augmented abstractions for planning that focus on single-robot object search~\cite{kulich_2014_object_search, schmalstieg2022learninglonghorizonrobotexploration, wani2020multionbenchmarkingsemanticmap, khanal2024dynamics,hossain2024enhancing} and task planning~\cite{bradley2021learning} in partially known environments, where a learned estimator makes predictions about unseen space needed to act well.
The planner of Khanal et al.~\cite{mrlsp} addresses multi-robot point-goal navigation, but their work cannot straightforwardly handle complex uncertain state transitions needed for coordination at the task planning level.
As such, we need an approach that both coordinates multiple robots to plan for complex tasks and uses learning to make predictions about unseen objects needed to inform effective search.

In this work, we develop a novel framework for multi-robot task planning under uncertainty and demonstrate effective multi-robot coordination in various task planning settings.
The central contribution of this work is a multi-robot planning framework that jointly achieves all of \circled{1}, \circled{2}, and \circled{3} to achieve coordinated long-horizon planning under uncertainty, subsuming prior work in this space~\cite{bradley2021learning, mrlsp, stein2018learning}.
Specifically, our framework introduces a multi-robot planning abstraction that incorporates learned predictions about object locations and accounts for complex state transitions that arise when multiple robots simultaneously execute stochastic, high-level actions with
non-uniform durations.

We demonstrate the efficient multi-stage task planning of our approach with improvements of up to 47.0\%, 40.7\%, and 33.8\% for 1, 2, and 3 robot teams over baselines in large \gls{ProcTHOR} household environments.
Additionally, we demonstrate the effectiveness of our approach with a team of two LoCoBot mobile robots in real household settings.

\section{Related Work}
Task planning has been widely explored in robotics, where the goal is to generate action sequences that enable robots to complete multi-stage objectives.
Classical symbolic planners often assume full knowledge of the environment \cite{McDermott1998PDDLthePD,zhang2025lamma}, which limits their applicability in realistic household settings where task-relevant objects may not be directly observable.

Executing tasks in unknown or partially known environments requires robots to search for and then interact with objects to make progress in the task.
Learning has proven to be a useful tool for informing good behavior in such settings and many works leverage learning to search for objects, ranging from heuristic-based exploration strategies to approaches that exploit semantic and structural cues to predict likely object locations~\cite{kulich_2014_object_search, schmalstieg2022learninglonghorizonrobotexploration, wani2020multionbenchmarkingsemanticmap, khanal2024dynamics,hossain2024enhancing} in previously unseen environments.
However, these approaches are limited to object search and not easily applied to solve complex multi-stage tasks.
Other works~\cite{bradley2021learning, littman1995learning,arnob2026effective} use temporal logic or \gls{PDDL} to specify tasks and leverage learning to plan for multi-stage tasks.
However, these approaches are designed for a single robot and cannot account for the concurrent, durative actions required to coordinate multi-robot teams.

Multiple robots working as a team can make concurrent progress towards a task to complete it more quickly.
Some recent works~\cite{rl1, marl, rl2} rely on model-free reinforcement-learning-based approaches for multi-robot coordinated planning under uncertainty.
While these approaches have shown promising results, they often require extensive training data and can be brittle~\cite{RLsurvey}, particularly for complex and long-horizon tasks.
Khanal et al.~\cite{mrlsp, khanal2025learning} use learning with model-based planning to coordinate a team of robots in an unknown environment to find unseen point goals.
However, this work is limited to goal-directed navigation and cannot be used to coordinate a team of robots to do task planning.

\vspace{0.4em}\noindent{}%
While prior works have made progress in using learning under uncertainty, task planning, and multi-robot coordination, these advances have largely been achieved in isolation.
There is a need for a representation that enables a team of robots to coordinate and plan multi-stage tasks under uncertainty regarding task-relevant object locations.
This work addresses this gap by leveraging learning to enable multi-robot teams to plan efficiently for complex tasks in partially known environments by reasoning about where objects may be, and using a model-based planning framework to capture how individual actions contribute to overall task progress and how to coordinate concurrent execution.

\begin{figure*}[t]
    \centering
    \vspace{0.5em}
    \includegraphics[width=\linewidth]{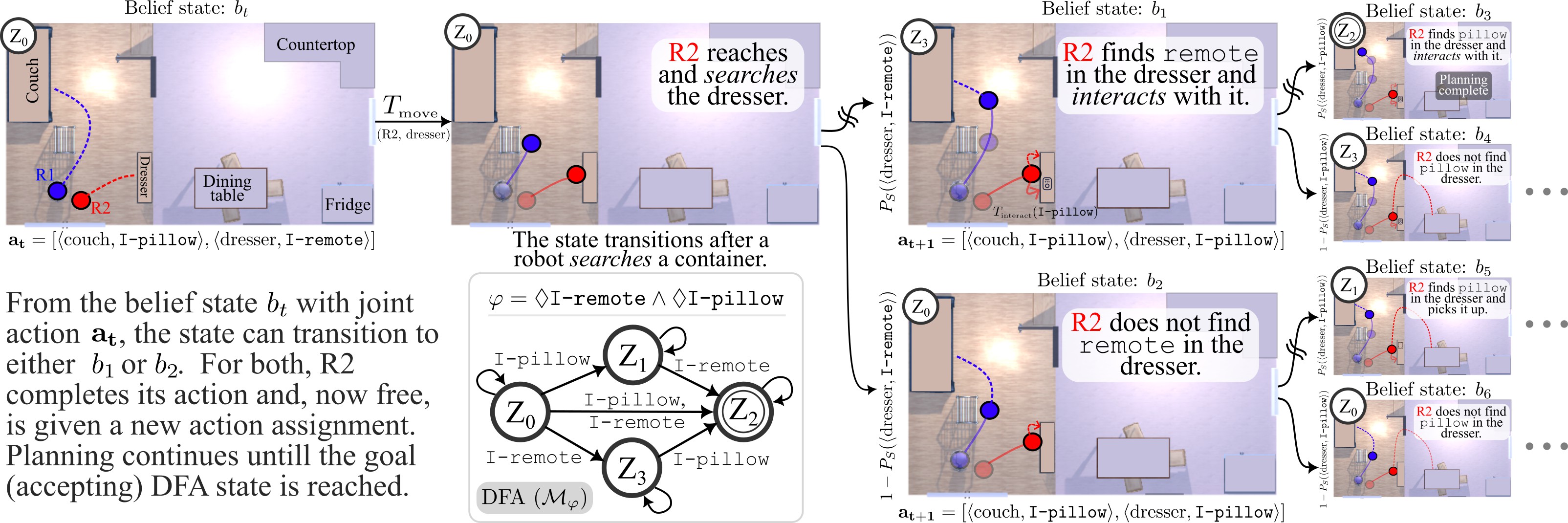}
    \vspace{-1.8em}
    \caption{\textbf{Overview of our approach:} For a joint-action $a_t$ specifying what container each robot should travel towards and interact with to find task-relevant objects, the robot team concurrently travels towards the assigned containers---until one of the robots reaches and \emph{searches} a container. The outcomes of each joint action transition the robot's belief state based on whether any task-relevant objects were found on the container. The \gls{DFA} $\mathcal{M}_\varphi$ keeps track of the overall task progress and transitions based on what objects have been interacted with and what objects are still needed to complete the task.}
    \label{fig:overview}
    \vspace{-1.8em}
\end{figure*}

\section{Preliminaries: Representing Tasks and Montioring Progress with \glsxtrshort{DFA} and \glsxtrshort{SCLTL}} \label{sec:LTLandDFA}
To enable effective coordination and execution for a multi-robot team, we require a task representation that allows us to reason about the team's progress, identify when the task is complete, and determine which actions will advance the task state.
A \gls{DFA} is an ideal representation for this purpose.
The states of the \gls{DFA} can represent the progress made, its transitions can map to robot actions that advance the task, and its accepting states formally define task completion.

While a \gls{DFA} provides the properties we need, manually constructing one for every complex task is impractical.
Instead, we specify tasks using \gls{SCLTL}.
\Gls{SCLTL} is a natural choice in this domain, owing to its expressive sytax for writing specifications with boolean and temporal operators---such as negation $(\neg)$, conjunction $(\land)$, eventually $(\Diamond)$, until $(\mathcal{U})$---over a set of atomic propositions $\Sigma$, each corresponding to a robot action.
Crucially, robust tooling exists to convert an \gls{SCLTL} formula into the desired \gls{DFA} representation~\cite{spot}.

For instance, consider a task where a robot team must pick both a \texttt{pillow} and a \texttt{remote} without constraining the order. We first define the atomic propositions as $\Sigma = \{\texttt{pick-pillow}, \texttt{pick-remote}\}$ and write the task in \gls{SCLTL} as:
$
 \varphi = \Diamond  \texttt{pick-remote} \land \Diamond \texttt{pick-pillow}
$.
This \gls{SCLTL}-specified task $\varphi$ can be converted into a $\gls{DFA}$, denoted as $\mathcal{M}\varphi = \langle \mathcal{Z}, Z_0, \Sigma, F \rangle$, used to monitor the progress towards completing the task.
Here, $\mathcal{Z}$ is the finite set of \gls{DFA} states with each $z \in \mathcal{Z}$ corresponding progress in the task, $Z_0 \in \mathcal{Z}$ is the initial state, $F \subseteq \mathcal{Z}$ is a set of accepting (goal) states.
The \glsxtrshort{DFA} induces a directed graph (as visualized in~\cref{fig:overview}) that specifies how the team can advance the state of the task.
The figure shows the initial state $Z_0$, where neither object has been picked: e.g., relevant words $\mathcal{W} \in \Sigma$ needed to advance the \glsxtrshort{DFA} is $\mathcal{W}(Z_0) = \{\texttt{pick-pillow}, \texttt{pick-remote}\}$.
If the team picks the \texttt{pillow} first, the \glsxtrshort{DFA} transitions to state $Z_1$, with $\mathcal{W}(Z_1) = \{\texttt{pick-remote}\}$.
Conversely, picking the \texttt{remote} first leads to state $Z_3$, with $\mathcal{W}(Z_3) = \{\texttt{pick-pillow}\}$.
The overall task is completed when an accepting state $Z_2$ is reached, where both pillow and remote have been picked.

\section{Problem Formulation}
We consider a team of $N$ centrally coordinated homogenous robots operating in a partially known environment. The environment is described by an occupancy grid $m_t$ and a set of known locations called \emph{containers} $\mathcal{S}$: e.g., tables, desks, beds.
The containers may contain objects (e.g., books, pillows, remotes), but the robot might not know which container contains which object at the outset.
Each robot has a set of \textit{interaction} (manipulation) skills like pick and place.
The team's high-level task is specified by an \gls{SCLTL} formula $\varphi$.
Each atomic proposition $\Sigma$ that forms the \gls{SCLTL} formula $\varphi$ corresponds to the execution of a robot skill on an a object: e.g, \texttt{pick-remote}, \texttt{reach-pillow}.

Since the location of task-relevant objects may be unknown to the robots, the robots have to search for task-relevant objects and then \emph{interact} with them to make progress towards the task.
Therefore, in addition to manipulation skills to \emph{interact} with objects, the robots also has skills to \emph{move} to and when necessary \emph{search} containers for objects.
Using the combination of all of these skills, necessary to complete the objective, the robot team has to coordinate to satisfy the task specification $\varphi$ in minimum time.

\begin{figure*}[t]
    \vspace{1em}
    \centering
    \footnotesize
    \begin{tabular}{lccc|ccc|ccc}
\toprule
                                      & \multicolumn{3}{c|}{\tikz \fill[myblue] (0,0) circle (2.5pt); Small}     & \multicolumn{3}{c|}{\tikz \fill[myred] (0,0) circle (2.5pt); Medium}    & \multicolumn{3}{c}{\tikz \fill[mygreen] (0,0) circle (2.5pt); Large}     \\
                                      & 1 robot & 2 robots & 3 robots & 1 robot & 2 robots & 3 robots & 1 robot & 2 robots & 3 robots \\
\midrule
Non-learned Myopic                    & 312.86  & 204.62   & 165.42    & 789.37  & 483.01   & 398.27    & 1413.62 & 857.10   & 649.77    \\
Learned Myopic                        & 255.34  & 160.03   & 133.03    & 672.05  & 368.57   & 286.64    & 1190.14 & 674.72   & 511.69    \\
MR Planner (ours)           & \textbf{215.22}  & \textbf{145.67}   & \textbf{130.86}    & \textbf{480.30}  & \textbf{304.11}   & \textbf{269.45}    & \textbf{748.37}  & \textbf{507.93}   & \textbf{429.83}    \\
\midrule
Improvement vs Non-learned Myopic    & 31.21\% & 28.81\%  & 20.89\%   & 39.15\% & 37.04\%  & 32.35\%   & 47.06\% & 40.74\%  & 33.85\%   \\
Improvement vs Learned Myopic        & 15.71\% & 8.97\%   & 1.63\%    & 28.53\% & 17.49\%  & 6.00\%    & 37.12\% & 24.72\%  & 16.00\%   \\
\bottomrule \\
\vspace{-2.1em}
\end{tabular}
\includegraphics[width=\linewidth]{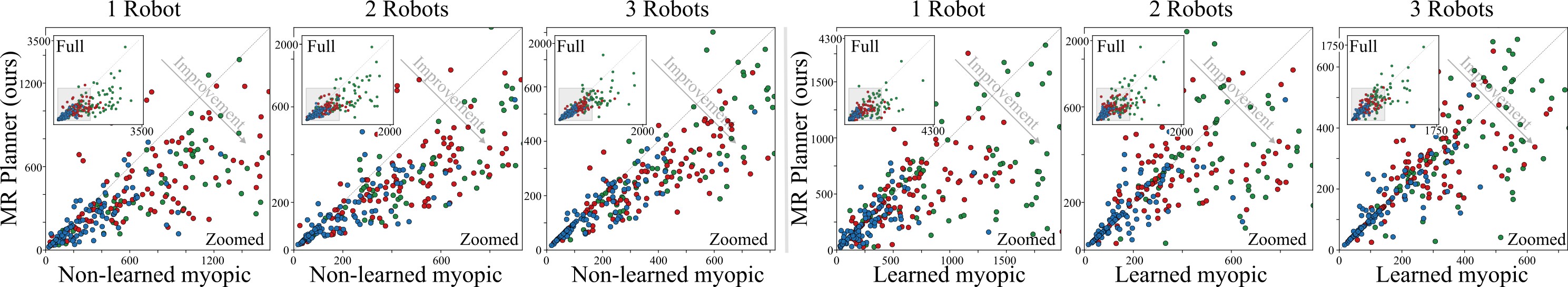}
\vspace{-2.2em}
    \caption{\textbf{Results: Partially known \Gls{ProcTHOR} environments}\quad The table shows the average cost accrued in 400 experiments in randomly selected tasks for each planner.
    We see that our planner improves over the cost over learned and non-learned baselines.
    Scatter plots shows results for our approach versus learned and non-leared baselines for 1, 2, and 3 robots. The statistics for all the results show benefits of our learning augmented model-based planner.}
    \vspace{-2.5em}
    \label{fig:scatter_result_partially_known}
\end{figure*}

\section{Methodology}
To enable the robot team to quickly complete complex long-horizon tasks under uncertainty, we first define a high-level abstract action for a single robot $r_i$. A single-robot high-level action $a_i \equiv \left<\sigma_i, w\right>$ corresponds to executing a sequence of skills: \textit{move} to container $\sigma_i \in \mathcal{S}$ (e.g., a couch), \textit{search} the container if necessary, and finally execute the \textit{interact} skill $w$ if its corresponding object is found in that container.
Via this action abstraction, the set of high-level abstract actions available to a single robot $\mathcal{A}_i$ is the outer product of the set of containers $\mathcal{S}$ and the set of interaction skills $w \in \mathcal{W}(Z)$ capable of advancing the \gls{DFA} state $Z$: $\mathcal{A}_i \equiv \mathcal{S} \otimes \mathcal{W}(Z)$.

To represent multiple robots concurrently making progress towards the task, a \emph{joint-action} representation $\mathbf{a_t}$ is needed.
The joint-action is written in terms of the individual single-robot actions assigned to each robot in the team: i.e., $\mathbf{a_t} = [\left<\sigma_1, w_1\right>, \left<\sigma_2, w_2\right>, \cdots, \left<\sigma_N, w_N\right>]$, where $N$ is the number of robots, so that $\mathbf{a_t} \in {\mathcal{A}_i}^N$.
Planning directly via this abstraction is difficult, since robots are acting concurrently and each of their actions may finish at different times and the duration of each action is conditioned on whether or not \emph{search} succeeds at finding the object of interest, necessitating a novel planning abstraction capable of tracking different ways the team might progress.

\subsection{A High-Level Abstraction for Multi-Robot Planning}
To plan, we seek to advance the state until a robot is \emph{free} and therefore in need of a new action. However, not only does each action have a non-uniform duration, but when a robot will become free depends on whether it succeeds in finding the object it is searching for in the container. Therefore, we need a state abstraction that tracks the progress of each robot towards their respective actions and determines all possible pathways for any robot to become available.

We define an abstract belief state as a tuple $b_t = \langle m_t, \mathcal{S}, \mathbf{q_t}, \mathbf{a_t}, \mathbf{p}, \mathcal{M}_\varphi,   Z, \mathcal{H} \rangle$, where $m_t$ is the occupancy map, $\mathcal{S}$ is the set of containers, $\mathbf{q_t}$ is the list of $N$ robot poses, $\mathbf{a_t}$ is the list of high-level action assigned to the robots with progress $\mathbf{p}$ the robots have made towards their assigned action,  $\mathcal{M}_\varphi$ is the \gls{DFA}, $Z$ is the current \gls{DFA} state, and $\mathcal{H}$ is the \emph{observation history}, which stores the outcomes of \emph{search}.
Whenever any robots are free, they are assigned an action from $\mathcal{A}(b_t) \equiv \mathcal{A}_i \cup \{\textit{wait}\}$, where the additional \textit{wait} action tells that robot to wait in place until another robot is free, sometimes needed for good behavior.

Planning must consider the many possible outcomes of search for a particular team assignment, how likely those outcomes are, and, conditioned on each, when each robot will complete its assigned action and so be in need of a new assignment.
For example, if there were only one robot executing an action $a_i$ to \texttt{pick-block} in the cabinet, two belief states are reachable: (1) with probability $P_S(a_i \equiv \langle\text{cabinet}, \texttt{pick-block}\rangle)$ the block is found and time passes until its \texttt{pick-block} interaction finishes; (2) with the inverse probability the block is not found, and the robot becomes free immediately after search reveals the block is not in the cabinet.
For the multi-robot case, each robot is concurrently moving, searching, or interacting, and so there are potentially many pathways to any of the robots becoming free, necessitating an approach to determine the distribution of belief states in which a robot needs assignment: $\mathcal{B}_\text{\tiny{}free}$.
From each of the resulting belief states, any free robots can then be assigned new actions, $\mathcal{A}(b) : b \in \mathcal{B}_\text{\tiny{}free}$.
In the team’s joint action, only the robots that are free get assigned a new action, while the actions of all other robots remain unchanged.

Algorithm \ref{alg:advance_belief} (\textsc{AdvanceBeliefUntilRobotFree}) details our procedure to compute the distribution $\mathcal{B}_\text{\tiny{}free}$.
This process works as follows: we first imagine the most optimistic outcome, in which each robot succeeds in finding the object they are looking for and can interact with it. The first of those actions to finish provides an upper bound on when a robot can become free. During that time, any of a number of search actions may fail to find the target object, and the relevant robot is then in need of a new action assignment. These outcomes---the single successful case and the possible negative search outcomes along the way---form the set of states that constitute $\mathcal{B}_\text{\tiny{}free}$.
Since each state $b'$ computed along the way contains observation history $\mathcal{H}_{b'}$ of past actions and outcomes, the probability $p_{b'}$ of reaching state $b'$ can be computed as as:
\begin{equation*} \label{eq:pb-prime}
p_{b'} = \prod_{(a, o) \in \mathcal{H}'}
\begin{cases}
P_S(a) & \text{if } o = \textsc{ObjectFound},\\
1 - P_S(a) & \text{if } o = \textsc{ObjectNotFound}
\end{cases}
\end{equation*}
Here, the likelihood $P_S(a)$ is estimated via supervised learning. The probability is normalized across all states to get a distribution over states $\mathcal{B}_\text{\tiny{free}}$.

\newcommand\mycommfont[1]{\scriptsize\textcolor{gray}{#1}}
\SetCommentSty{mycommfont}
\SetKwComment{Comment}{// }{}
\SetKw{Break}{break}
\SetAlgoNoEnd
\providecommand{\nonl}{\renewcommand{\nl}{\relax}}
\makeatletter
\renewcommand*{\@algocf@post@ruled}{
  \vspace{0.1em}
  \noindent\rule{\linewidth}{0.9pt}
  \begin{minipage}{\linewidth}
    \vspace{0.4em}
    \includegraphics[width=0.97\linewidth]{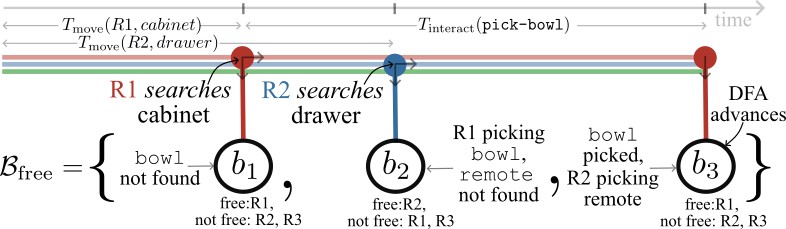}
  \end{minipage}
\vspace{-2.0em}
}
\makeatother
\begin{algorithm}[t]
\label{alg:advance_belief}
\LinesNumbered
\DontPrintSemicolon
\footnotesize
\caption{\textsc{AdvanceBeliefUntilRobotFree}}
\KwIn{$b_t = \langle m_t, \mathcal{S}, \mathbf{q_t}, \mathbf{a_t}, \mathbf{p}, \mathcal{M}_\varphi, Z_0, \mathcal{H} \rangle$}
\KwOut{$\mathcal{B}_\text{\tiny{}free}$}

$\mathcal{T}_{\text{S}} \gets
  \{ (a,\; T_{\text{move}}(r,a.\sigma)
     + T_{\text{interact}}(w))
     \;\mid\; (r,a) \in \mathbf{a_t}, a \neq \text{wait} \}$\;

$\mathcal{T}_{\text{F}} \gets \{ (a, T_{\text{move}}(r,a.\sigma)) \mid (r,a) \in \mathbf{a_t} \}$\;
\Comment{Sort action outcomes (success or failure) by time of completion}
$\mathcal{T} \gets \textsc{SortByTime}(\mathcal{T}_{\text{S}} \cup \mathcal{T}_{\text{F}})$\;
$\mathcal{B}_\text{\tiny{}free} \gets \emptyset$, $\mathcal{H}' \gets b_t.\mathcal{H}$\;
\ForEach{$(a,t) \in \mathcal{T}$}{
  $\mathbf{q'}, \mathbf{p'} \gets \textsc{ForwardSimulateRobots}(b_t,t)$\;
  \eIf{$(a,t) \in \mathcal{T}_{\text{F}}$}{
     \Comment{Compute negative search outcomes along the way}
     $\mathcal{H'} \gets \mathcal{H}' \cup \{(a,\textsc{ObjectNotFound})\}$\;
     $b' \gets \langle m_t, \mathcal{S}, \mathbf{q'}, \mathbf{a_t}, \mathbf{p'}, \mathcal{M}_\varphi, Z_0, \mathcal{H'} \rangle$\;
     $p \gets \textsc{ComputeProbabilityFromHistory}(\mathcal{H'})$\;
     $\mathcal{B}_\text{\tiny{}free} \gets \mathcal{B}_\text{\tiny{}free} \cup \{ b' : (p,t) \}$\;
     \Comment{Object is found for the next state}
     $\mathcal{H}' \gets b_t.\mathcal{H} \cup \{(a,\textsc{ObjectFound})\}$\;
  }{
    \Break\;
  }
}
\Comment{Compute most optimistic state}

$\mathcal{H}' \gets \mathcal{H}' \cup \{(a,\textsc{ObjectFound})\}$\;
$\mathcal{M}_\varphi' \gets \mathcal{M}_\varphi.\textsc{AdvanceDFA}(a.w)$\;
$b' \gets \langle m_t, \mathcal{S}, \mathbf{q'}, \mathbf{a_t}, \mathbf{p'}, \mathcal{M}_\varphi', \mathcal{M}_\varphi'.\text{state}, \mathcal{H}' \rangle$\;
$p \gets \textsc{ComputeProbabilityFromHistory}(\mathcal{H'})$\;
$\mathcal{B}_\text{\tiny{}free} \gets \mathcal{B}_\text{\tiny{}free} \cup \{ b' : (p,t) \}$\;
$\mathcal{B}_\text{\tiny{}free} \gets \textsc{NormalizeProbability}(\mathcal{B}_\text{\tiny{}free})$\;
\Return{$\mathcal{B}_\text{\tiny{}free}$}; \Comment{\{$b_1$: ($P_{b_1}$, cost), $b_2$: ($P_{b_2}$, cost) $\cdots $\}}
\nonl
\end{algorithm}

\subsection{Planning with our Multi-Robot Abstraction}
Given the belief state $b_t$, and a joint-action $\mathbf{a_t}$, we can compute distributions over states $\mathcal{B}_{\tiny{\text{free}}}$, where robots are free.
From these states, the robot team plans until the team reaches the accepting (goal) \gls{DFA} state, where the task is complete.
This whole process is illustrated in in~\cref{fig:overview}, and the cost of a multi-level joint action is defined by a Bellman Equation:
\begin{equation} \label{eq:mr-task-recursion}
Q(b_t, \mathbf{a_t}) = \!\!\!\! \sum_{b' \in B_\text{free}} \!\!\!\! P_{b'} \Big[t_{b'} + \!\! \min_\mathbf{a_t'\in \mathcal{A}(b')}Q(b', \mathbf{a_t'})\Big]
\end{equation}
Here, $P_{b'}$ is the likelihood of reaching the state $b'$ from state $b_t$ under joint action $\mathbf{a_t}$.

\subsection{Learning to Estimate Probabilisitic Action Outcomes} \label{sec:learning_state_transition}
The state that the robot can reach after executing an action $a_i\equiv\langle\sigma_i, w_i\rangle$ depends on the outcome of the action.
For instance, if the action $ a \equiv \langle\text{cabinet}, \texttt{pick-block}\rangle$, i.e., \emph{move} to the cabinet and execute skill \texttt{pick-block}, the outcome depends on whether or not the block is on the cabinet.
So, we train a neural network to estimate $P_S(a_i\equiv\langle\sigma_i, w_i\rangle)$, the likelihood that the task-relevant object for skill $w_i$ is present on the container $\sigma_i$.
We discuss the details of our neural network to learn the likelihood in~\cref{sec:neural-network}.
By estimating this likelihood via learning, we can compute the transition probabilities of different outcomes in $\mathcal{B}_{\text{\tiny{free}}}$.

\subsection{Computing Planning Costs for the Team using PO-UCT}
We compute the cost of multi-robot joint-action using a sample-efficient any-time planning strategy via Partially Observable \textsc{Uct} (\textsc{po-uct}) \cite{silver2010monte}, a variant of Monte Carlo Tree Search.
This allows us to approximate expected costs through selective sampling rather than exhaustive simulation.

Planning is performed by incrementally building a search tree rooted at the current belief state $(b_t)$.
Each node in this tree maintains a rollout history $\mathcal{RH}_{b_t} = [[a_0,n_0,Q_0],\cdots,[a_k,n_k,Q_k]]$, which stores high-level actions $a_i \in \mathcal{A}(b_t)$, the count $n_t$ for each action, and the accumulated cost estimates $Q_t$ associated for each action $a_i$.

Tree expansion is driven from the belief state $b_t$, where robots are assigned a high-level action from the set $\mathcal{A}(b_t)$ and a new node is created.
From the node where all robots are assigned forming a multi-robot joint-action, the resulting node is sampled using a Bernoulli trial from the distribution over states $\mathcal{B}_\text{\tiny{free}}$, computed using Algorithm~\ref{alg:advance_belief} where one or more robots are free.
From the new node's belief state, free robots are assigned actions again, and the planning continues.

The rollout mechanism closely resembles that of standard \textsc{mcts}, with each node's cost estimated as the sum of (i) the cost to reach that belief from the root and (ii) a heuristic cost providing a lower bound on the remaining cost to the goal.
After each node is visited, its visit count is incremented to balance exploration and exploitation during search. Each planning iteration uses $10^5$ samples to guide decision making.
During execution, the team selects a joint action with minimum cost and executes it.
The team replans once one robot finishes executing its action and becomes free.

\begin{figure}[t]
    \centering
    \vspace{0.5em}
    \includegraphics[width=\linewidth]{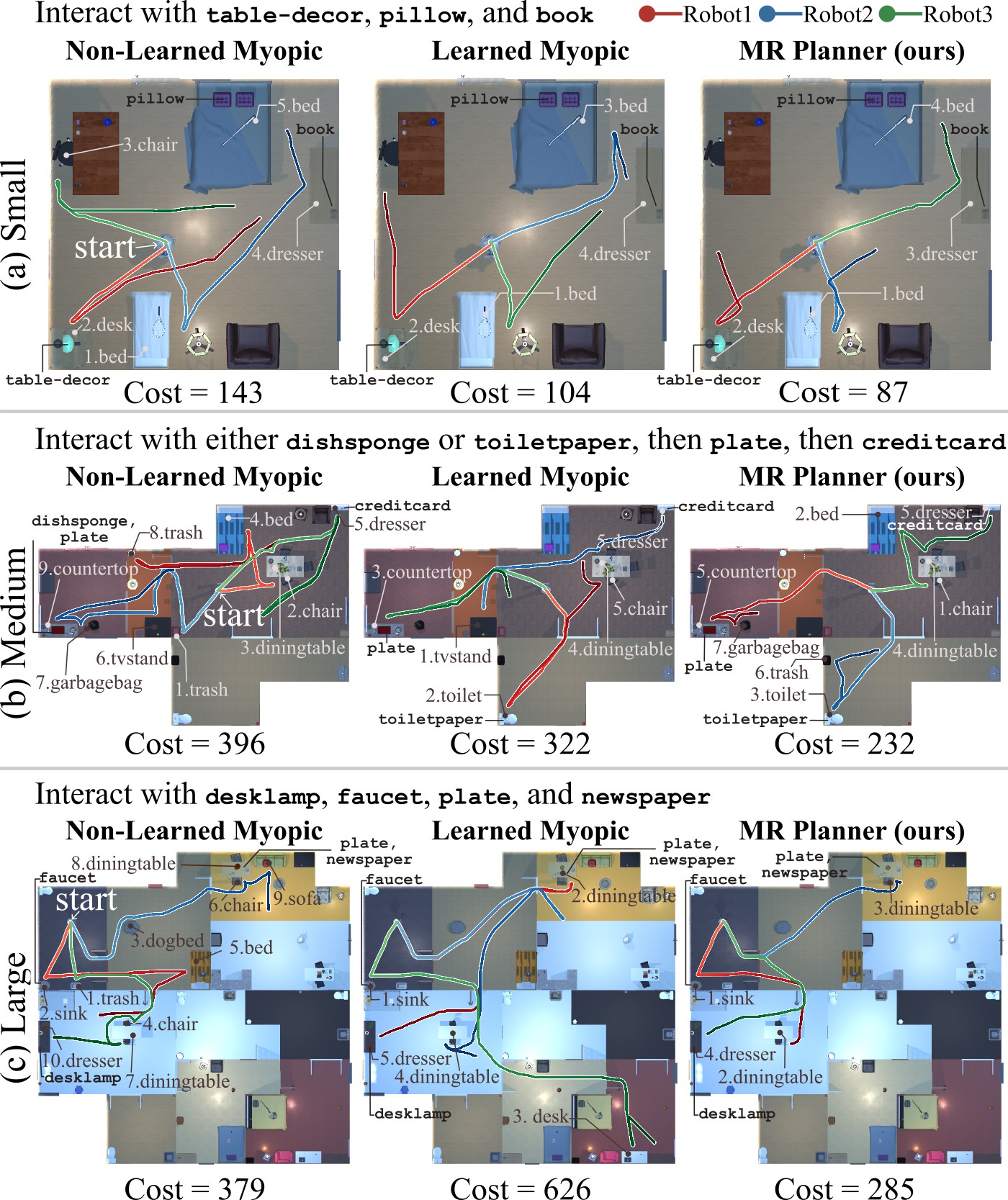}
    \vspace{-1.6em}
    \caption{\textbf{Simulation experiments} in small, medium, and large environments for 3 robots.
    (a) Task: Interact with a \texttt{pillow}, a \texttt{tabletopdecor}, and a \texttt{book}.
    (b) Task: Interact with either a \texttt{dishsponge} or a \texttt{toiletpaper}, then interact with a \texttt{plate}, then interact with a \texttt{creditcard}.
    (c) Task: Interact with a \texttt{desklamp}, a \texttt{faucet}, a \texttt{plate}, and a \texttt{newspaper}.
    In all these tasks, across different environment sizes, our MR planner improves cost over non-learned and learned baselines.}
    \vspace{-1.5em}
    \label{fig:small_medium_big}
\end{figure}

\section{Simulation Experiments} \label{sec:procthor_experiments}
\begin{figure}[t]
    \centering
    \vspace{0.5em}
    \includegraphics[width=\linewidth]{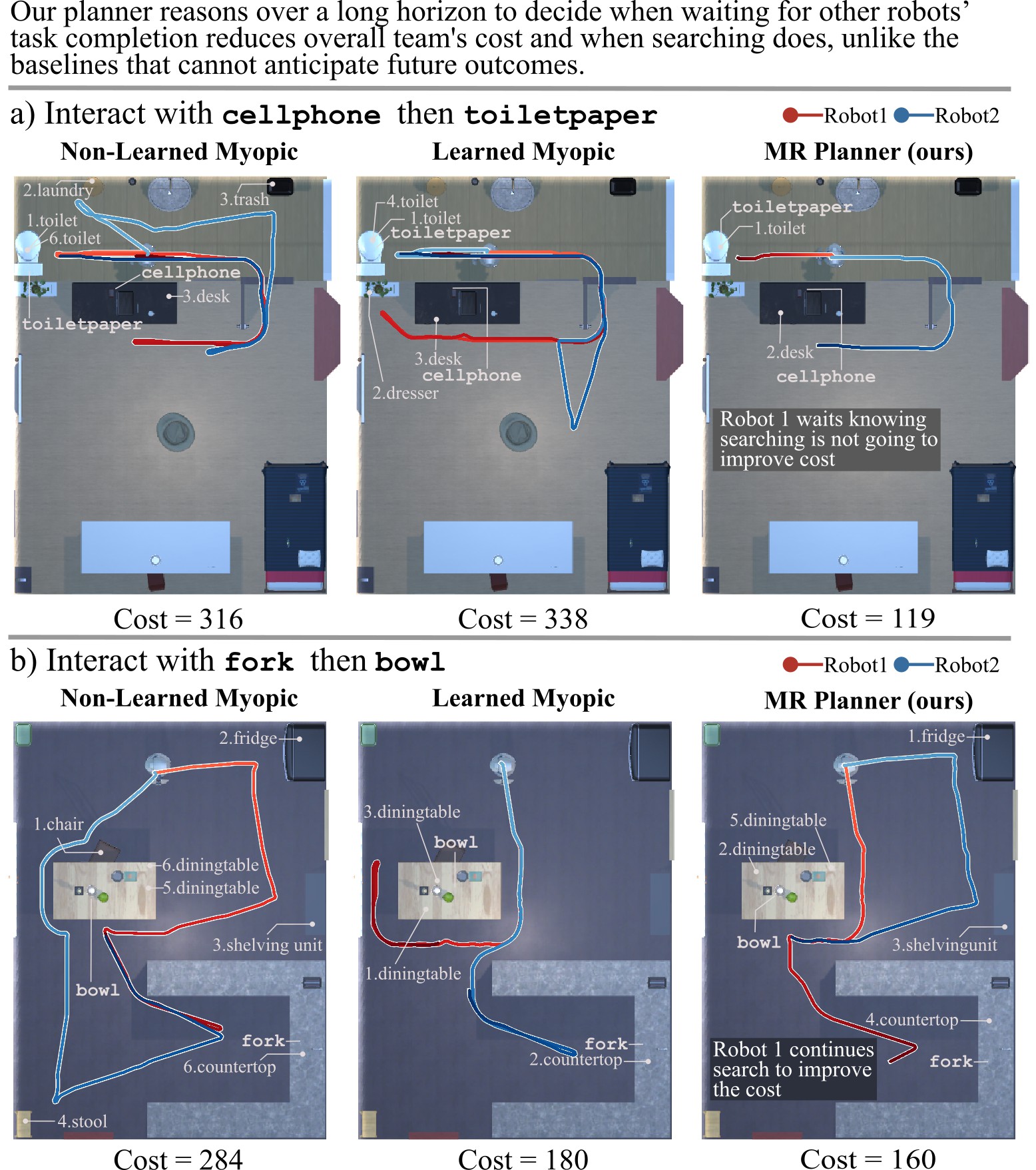}
    \vspace{-1.9em}
    \caption{Robots are tasked with order-dependent tasks with strict temporal constraints: (a) interacting with \texttt{cellphone} then \texttt{toiletpaper}, and (b) interacting with \texttt{fork} then \texttt{bowl}. Robots in our model-based planner (a) wait to interact with the object when searching for remaining objects does not improve task-completion cost, and (b) continue to search for other task-relevant objects when waiting to interact with the found object would increase the task-completion cost. The baselines search for all task-relevant objects and then resolve temporal dependencies of the task.}
    \vspace{-1.5em}
    \label{fig:waiting}
\end{figure}

\begin{figure}[t]
    \centering
    \vspace{0.5em}
    \includegraphics[width=\linewidth]{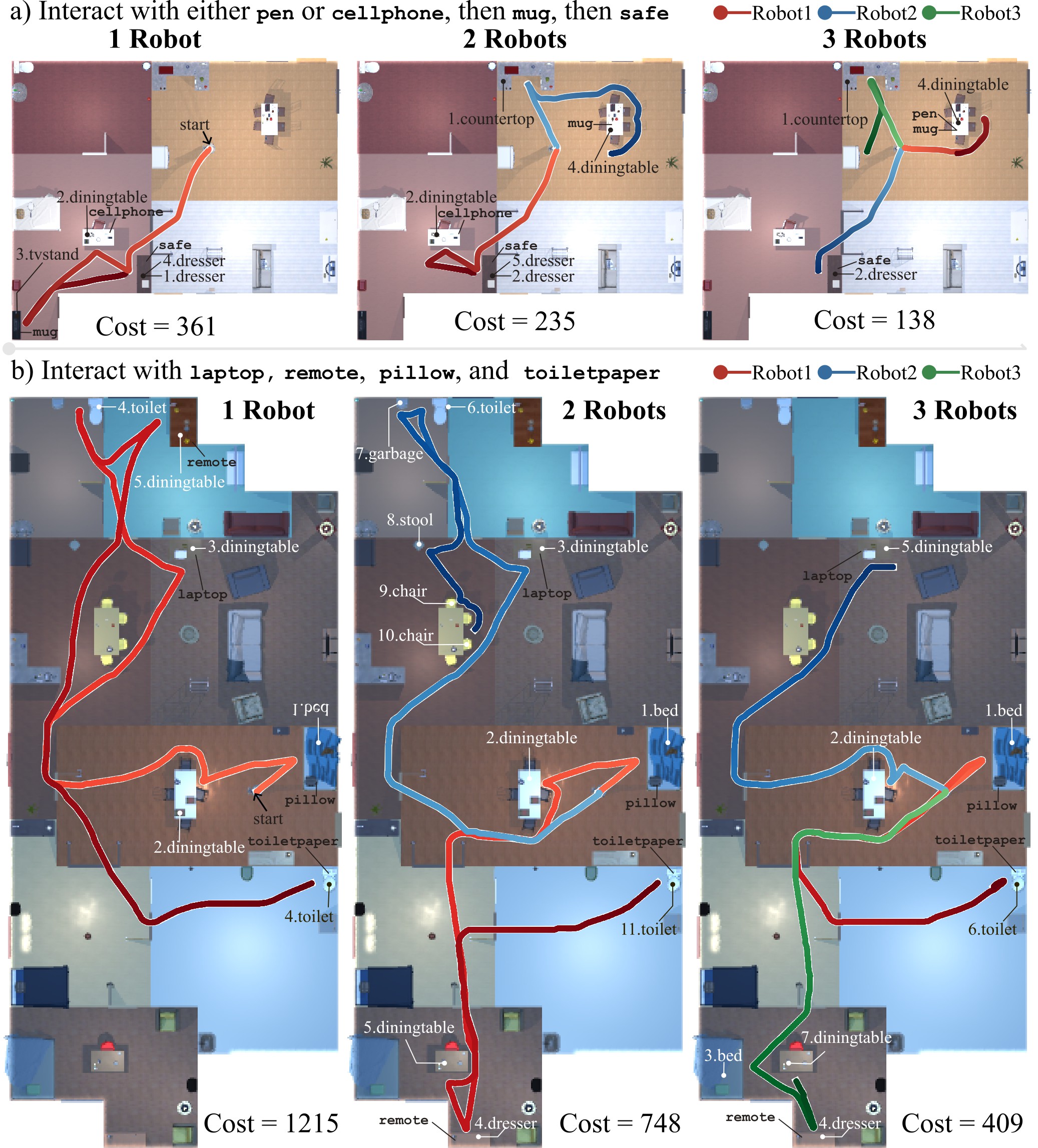}
    \vspace{-1.8em}
    \caption{As the number of robots increase, our planner coordinates robots effectively, divides efforts between the team to improve cost for both (a) complex order-dependent tasks and (b) order-independent tasks.}
    \vspace{-1.8em}
    \label{fig:robots_comparison}
\end{figure}

We conduct simulated experiments in partially known, procedurally generated homes using \gls{ProcTHOR}~\cite{procthor}.
Each home contains a variety of \emph{containers}---cabinets, beds, tables, countertops---that may hold objects relevant to a given task specification: e.g., a phone, laptop, a remote, and others.
The locations of these containers are known to the robot, yet what objects are present in those containers are not known to the robot team at the outset.
As such, the team must coordinate to search these containers and find and grab the required objects in the order specified by the task specifications to complete the task.

\subsection{Task Specification Templates}
In order to generate a diverse set of experiments in these home environments, we procedurally create task specifications using templates that are then populated with objects from the environment.
Each task expects the team to find and interact with various objects across the environment.
To formalize this, we define \texttt{I-o} as an interaction with object \texttt{o}, where interaction corresponds to executing a skill (e.g., \emph{pick}, \emph{grasp}, \emph{reach}) on that object.
In experiments, we generate tasks at random from the following set of specifications:
\begin{itemize}[leftmargin=*]
    \item Interact with \texttt{a} and \texttt{b}: $\Diamond \texttt{I-a} \land \Diamond \texttt{I-b}$.

    \item Interact with \texttt{a}, \texttt{b}, and \texttt{c}: $\Diamond\texttt{I-a} \land  \Diamond\texttt{I-b} \land \Diamond\texttt{I-c}$.

    \item Interact with \texttt{a}, and either \texttt{b} or \texttt{c}: $\Diamond\texttt{I-a} \land  (\Diamond\texttt{I-b} \lor \Diamond\texttt{I-c}$)

    \item Interact with \texttt{a}, then with \texttt{b}: $(\neg \texttt{I-b} \ \mathcal{U} \ \texttt{I-a}) \land \Diamond \texttt{I-b}$.

    \item Interact with \texttt{a}, then interact with \texttt{b}, then interact with \texttt{c}: $((\neg \texttt{I-c} \land \neg \texttt{I-b})\ \mathcal{U}\  \texttt{I-a}) \land (\neg \texttt{I-c}\ \mathcal{U}\ \texttt{I-b}) \land \Diamond \texttt{I-c}$

    \item Interact with either \texttt{a} or \texttt{b}, then interact with \texttt{c}, then interact with \texttt{d} : $((\neg \texttt{I-c} \land \neg \texttt{I-d})\ \mathcal{U}\  \texttt{I-a}) \lor
((\neg \texttt{I-c} \land \neg \texttt{I-d})\ \mathcal{U}\ \texttt{I-b}) \land
              (\neg \texttt{I-d}\ \mathcal{U}\ \texttt{I-c}) \land \Diamond \texttt{I-d}$
\end{itemize}
Here, $\texttt{a}$, $\texttt{b}$, $\texttt{c}$, and $\texttt{d}$ are randomly sampled from objects present in the containers in any home.

The tasks fall into two broad categories.
The first includes order-independent tasks, where objects could be found and interacted with in any order (e.g., $\Diamond\texttt{I-a} \land \Diamond \texttt{I-b}$, or $\Diamond\texttt{I-a} \land  (\Diamond\texttt{I-b} \lor \Diamond\texttt{I-c}$)).
The second category includes order-dependent tasks with strict temporal constraints, such as $(\neg \texttt{I-a} \ \mathcal{U} \ \texttt{I-b}) \land \Diamond \texttt{I-a}$, which requires finding and interactions with \texttt{b} before interactions with \texttt{a}.
The combination of these can represent complex task sequences, requiring the robot team to coordinate to complete the tasks.
\begin{figure}[ht]
    \centering
    \vspace{0.5em}
    \includegraphics[width=\linewidth]{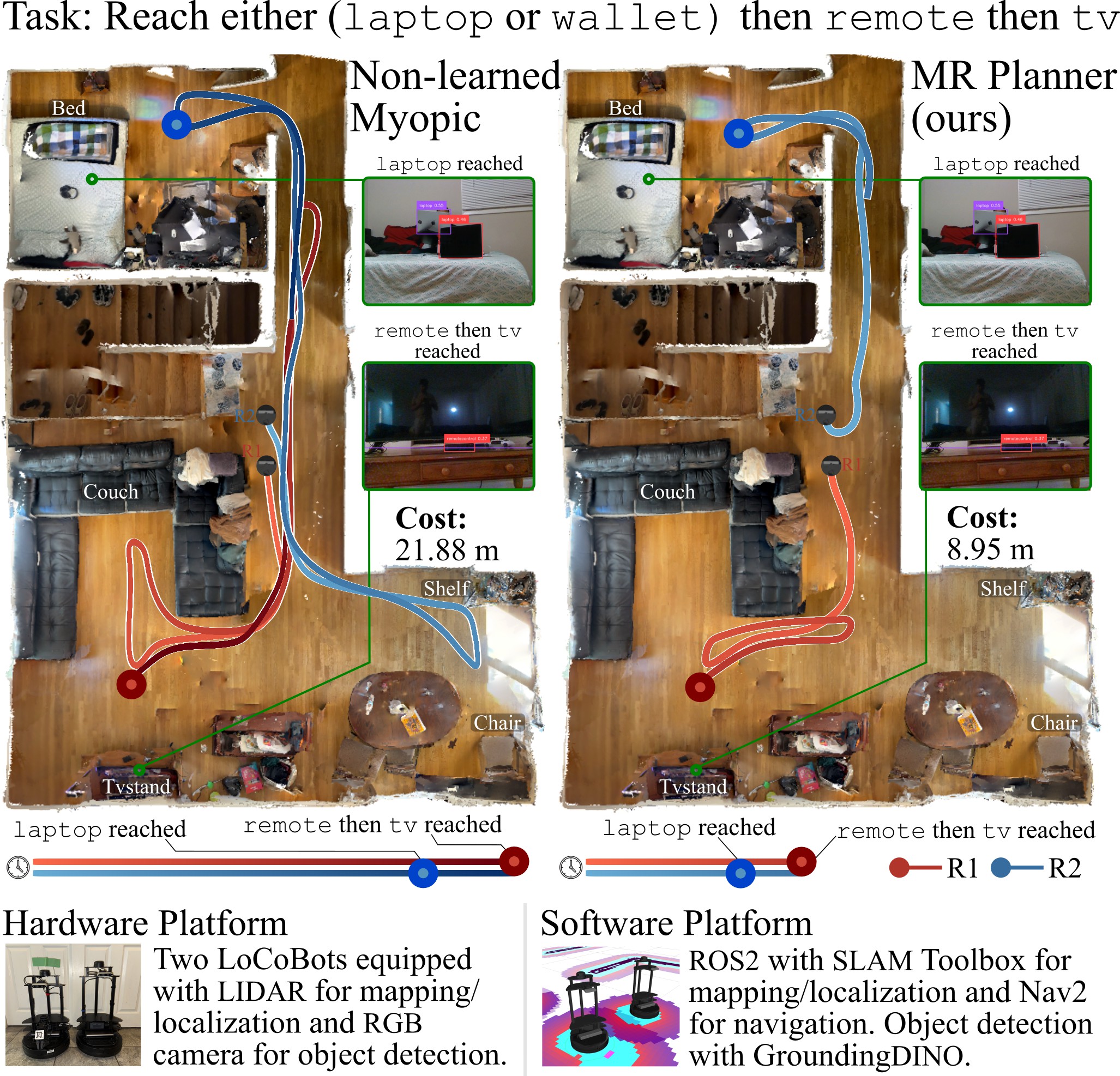}
    \vspace{-1.7em}
    \caption{\textbf{Real-world Demonstration:} For the task of reaching either (\texttt{laptop} or \texttt{wallet}) then \texttt{remote} then \texttt{tv} in a home with a two-robot team, our learning-informed model-based planner outperforms the non-learned myopic planner in terms of navigation costs.}
    \vspace{-1.7em}
    \label{fig:real_world_resuts}
\end{figure}

\subsection{Data Generation and Neural Network Training} \label{sec:neural-network}
As discussed in~\cref{sec:learning_state_transition}, we estimate the likelihood of task-relevant objects in the containers.
We use a neural network trained on a supervised dataset generated from 500 \gls{ProcTHOR} environments.
For each environment, we assign a positive label if the object is in the target container and a negative label otherwise.
The network takes as input \gls{SBERT} embeddings of the names of object, container, and room, each a 772-dimensional vector concatenated together, which is passed through 10 fully connected layers, with dimensions of 2048 to 4 in the decreasing powers of 2.
Each fully connected layer is followed by a batch normalization and a leaky \textsc{r}e\textsc{lu} activation.
Finally, an output layer with a sigmoid activation function predicts the desired likelihood, $P_S(\sigma,w)$.

\subsection{Planner Evaluation}
We run simulated trials with robot teams in procedurally generated homes using \gls{ProcTHOR}.
Each home is categorized into one of three size classes---small, medium, or large---based on its floor area.
Homes smaller than 60 $m^2$ are classified as small, those between 60 $m^2$ and 110 $m^2$ as medium, and those larger than $\textit{110 }m^2$ as large.
For each of these apartment sizes, we run 100 experiments for 1, 2, and 3 robots resulting in a total of 900 experiments, and evaluate the performance of the following planners:
\begin{description}[leftmargin=10pt]
    \item[\textbf{Non-learned Myopic Planner:}] This planner guides the robot team to search the nearest containers to interact with task-relevant objects.
    If the object is not found, the robot team replans to the next nearest containers until the task is completed (i.e., the planner reaches accepting \gls{DFA} state).
    \item[\textbf{Learned Myopic Planner:}] This planner guides the robots to explore the location with the highest likelihood of finding any of the objects that the team needs to eventually advance the \gls{DFA} state $Z$ in the task specification.
    For that, each container's $\sigma \in \mathcal{S}$ weight $P(\sigma)$ is computed by taking the sum of the likelihood of finding all task-relevant objects corresponding to $ w \in \mathcal{W}(Z)$ from the current \gls{DFA} state $Z$ using~\cref{eq:MR-task-learned-baseline}.
    \begin{equation}\label{eq:MR-task-learned-baseline}
    P(\sigma) = \sum_{\forall w: w \in \mathcal{W}(Z)} P_S(\sigma, w)
    \end{equation}
    Robots coordinate to minimize overall travel distance to those containers and re-plan until the task is complete.
    \item[\textbf{Multi-robot Model-Based Planner (Ours):}] This planner uses learned likelihood of task-relevant object locations with our model-based task planning to guide the team to complete the task specification.
\end{description}

\subsection{Results and Discussion}
We present the statistical results of all experiments in \cref{fig:scatter_result_partially_known}, which includes both a scatterplot and a summary table with average costs for all experiments.
Each point in the scatterplot corresponds to one experiment, with the cost incurred by our planner on the y-axis and the baseline cost on the x-axis.
Across all experiments with 1, 2, and 3 robots, our planner consistently outperforms both the non-learned and learned myopic baselines.
More importantly, this performance gap increases as the environment size increases.
Efficiently completing tasks in a large environment with multiple robots requires coordination and foresight to effectively split up the search effort between the team without redundant efforts.
Baseline approaches lack the long-horizon coordination, and as a result, perform increasingly poorly as the environment size grows.
In contrast, our approach leverages learning and plans by reasoning over a long horizon to effectively coordinate the team even in larger environments.

Representative trials in small, medium, and large environments, as shown in \cref{fig:small_medium_big}, further highlight the benefit of our planner's learning-informed long-horizon coordination.
Our approach effectively balances searching nearby locations with high-likely locations---for example, it plans by reasoning over when searching a nearby container for a less probable object first is more effective than traveling directly to a location that has a high probability of finding that object and potentially having to backtrack later.
This is evident in \cref{fig:small_medium_big}(a), where the planner directs the green robot to check a dresser for a \texttt{book} while en route to a bed, even though the bed is the more likely location. Similarly, in \cref{fig:small_medium_big}(c), the green robot searches a dresser for a \texttt{desklamp} on its way to the desk.
In contrast, the baseline planner assigns robots based purely on proximity without long-horizon reasoning and increases overall cost (e.g., in~\cref{fig:small_medium_big} (a), in the learned myopic baseline, the green robot skips the dresser en route to the bed and later returns to it to find the \texttt{book}).

We see various emergent coordination behaviors in the robot team when completing a complex task specification that requires satisfying a temporal dependency.
In~\cref{fig:small_medium_big}(b), the robots are tasked to interact with either \texttt{dishsponge} or \texttt{toiletpaper}, then a \texttt{plate}, and finally a \texttt{creditcard}.
Here, our planner assigns robots to relevant containers to find and interact with \texttt{toiletpaper}, \texttt{plate}, and \texttt{creditcard} one after another in that order, and handles dependencies efficiently through long-horizon planning and decision-making.
Behavior like this is mostly seen in medium and large environments with many containers, where robots have options to find the same objects in different containers.
Baselines often attempt to locate relevant objects and then try to make progress (requiring backtracking and interaction with he object when the task order is not satisfied), leading to inefficient paths.

Further examples are shown in \cref{fig:waiting}.
In (a), where the task is to interact with \texttt{cellphone} followed by \texttt{toiletpaper}, the red robot finds \texttt{toiletpaper} first but waits, anticipating that the blue robot will find and interact with the \texttt{cellphone}, thus minimizing task cost.
Similarly, in (b), where the task is to interact with \texttt{fork} then \texttt{bowl}, the red robot avoids prematurely interacting with \texttt{bowl} after finding it, instead continuing its search for \texttt{fork}, knowing that the blue robot will efficiently complete the interaction with \texttt{bowl} afterward.
The baselines do not anticipate how the action of robots might make progress towards the task in the future, hence replans if interaction with the object does not advance the \gls{DFA} state, accruing more cost.

We show that as the number of robots increases, the planner effectively splits efforts between the robot team and completes the task in minimum cost, showing effective coordination as shown in \cref{fig:robots_comparison}.
This improvement is also reflected quantitatively in the results table in \cref{fig:scatter_result_partially_known}.

Overall, while simple heuristics or likelihood-based reasoning can complete tasks, they often do so in more time hence incurring more cost.
By leveraging learning---to infer what any robot's action might reveal---with a model-based planning framework---for long-horizon coordinated decision making, our approach achieves multi-robot coordination and efficiently handles complex task specifications.

\section{Real-world Robot Experiments} \label{sec:real-robot-experiments}
We evaluated our multi-robot model-based task planning on a physical team of two LoCoBots (~\cref{fig:real_world_resuts}), each equipped with a \textsc{rgb} camera to recognize objects using GroundingDINO~\cite{liu2023grounding}.
We present experiments in two household environments (\cref{fig:intro,fig:real_world_resuts}).
In both of these environments, a static map was built from LoCoBot \textsc{lidar} scans, where container locations were then defined.
Robots are initially placed near each other at start locations and assigned a task to complete.
For predictions of object likelihoods in containers, we use the same neural network trained in \gls{ProcTHOR}.
We compare our learning-informed planner with a non-learned myopic baseline and show the results in \cref{fig:intro,fig:real_world_resuts}.

\Cref{fig:intro} shows results in a large apartment with living room, dining room, bedroom, and kitchen, and consisting of nine container locations: sidetable, bed, oven, drawer, countertop, couch, table, chair, and sofa.
These containers may hold the objects \texttt{pillow} and \texttt{remote}.
The task for the robot is to reach \texttt{pillow} and \texttt{remote} without any order.
The baseline myopic planner searches nearby containers first, leading both robots to explore containers in the kitchen before moving to the bedroom.
In bedroom, R1 finds the \texttt{pillow} on the bed, after which both robots search the living room, R1 finds the \texttt{remote} on the couch, completing the task with a navigation cost of 23.72 meters.
In contrast, our planner effectively coordinates the search and directs R1 to bedroom and R2 to livingroom.
In bedroom, R1 finds \texttt{pillow} at bed, and in living room, R2 robot finds \texttt{remote} at couch, completing the task with a navigation cost of 7.79 meters and resulting in 67.2\% improvement of  over baseline myopic planner in terms of navigation costs.

\Cref{fig:real_world_resuts} shows results in a medium apartment with living room, dining room, and bedroom with containers: chair, shelf, couch, tvstand, and bed.
The task for the robot is to reach either \texttt{laptop} or \texttt{wallet}, then \texttt{remote}, and then \texttt{tv} in that order.
While myopic planner shows poor coordination of robots, we observe that our planner, leveraging learning to predict likely object locations, effectively coordinates the robot team to search for the task-relevant objects and complete the task.
With our planner, R1 goes towards the living room and R2 moves towards the bedroom. In living room at tv stand, R1 finds \texttt{remote} and \texttt{tv}.
However, reaching \texttt{remote} or \texttt{tv} cannot currenlty result in task progress (i.e. advance the \gls{DFA}), and hence R1 continues to instead search for \texttt{laptop} or \texttt{wallet} elsewhere.
Later when R2 reaches the \texttt{laptop} at bed, R1 replans to come back to tv stand now that reaching \texttt{remote} and then \texttt{tv} can progress the task. After R1 reaches the \texttt{tv} and then the \texttt{remote}, the task is complete.
This results in the navigation cost of 8.95 meters for our planner while the myopic planner incurs 21.88 meters---an improvement of 59.1\%.

\section{Conclusion and Limitations}
In this work, we present a planning framework for multi-robot task planning under uncertainty that enables teams of robots to coordinate effectively in partially known environments.
Our approach integrates learning with model-based planning to reason about where task-relevant objects may be, how individual robots can concurrently execute actions and complete multi-stage tasks effectively.
Through experiments in simulated household environments and real-world deployments with mobile robots, we demonstrate effective team coordination to complete complex tasks and thus improve cost over non-learned and learned baselines.

While these results demonstrate learning-informed long-horizon coordination, there are still limitations to its current applicability.
This work is limited to a homogeneous robot team of having a similar skill set (e.g., pick, place).
In addition, general task planning often involves satisfying preconditions (e.g., the robot that picks an item must also be the one to place it).
Constructing \gls{SCLTL} specifications with such preconditions for general task planning quickly becomes difficult.
Furthermore, as with most centralized multi-robot planning frameworks, the joint search space grows with the number of robots and the complexity of the task specification, increasing computational demand.
All present opportunities for future work and exploration.

\section*{Acknowledgments}
This work was supported by the National Science Foundation (NSF) under Grant 2232733, and the U.S. Army Research Laboratory (ARL) under Grant W911NF2520011.

\bibliography{references.bib}
\bibliographystyle{IEEEtran}

\end{document}